\documentclass[letterpaper]{article} 
\usepackage{aaai25}  
\usepackage{times}  
\usepackage{helvet}  
\usepackage{courier}  
\usepackage[hyphens]{url}  
\usepackage{graphicx} 
\usepackage{natbib}  
\usepackage{caption} 
\usepackage{caption} 
\usepackage{amsfonts} 
\usepackage{algorithm}
\usepackage{algorithmic}
\usepackage{booktabs}
\usepackage{amsmath}

\usepackage{newfloat}
\usepackage{listings}
\DeclareCaptionStyle{ruled}{labelfont=normalfont,labelsep=colon,strut=off} 
\floatstyle{ruled}
\newfloat{listing}{tb}{lst}{}
\floatname{listing}{Listing}
\usepackage{enumitem}
\usepackage{tikz}
\usepackage{caption}
\usepackage{subcaption}
\usetikzlibrary{positioning, shapes.geometric, arrows.meta, calc, decorations.pathreplacing}

\title{EMERGENT: Efficient and Manipulation-resistant Matching using GFlowNets}
\author {
    Mayesha Tasnim,
    Erman Acar,
    Sennay Ghebreab
}
\affiliations {
    Socially Intelligent Artificial Systems, University of Amsterdam\\
    m.tasnim@uva.nl, e.acar@uva.nl, s.ghebreab@uva.nl
}

\usepackage{bibentry}

\begin{document}

\maketitle

\begin{abstract}
The design of fair and efficient algorithms for allocating public resources, such as school admissions, housing, or medical residency, has a profound social impact. 
In one-sided matching problems, where individuals are assigned to items based on ranked preferences, a fundamental trade-off exists between efficiency and strategyproofness.
%
%
Existing algorithms like Random Serial Dictatorship (RSD), Probabilistic Serial (PS), and Rank Minimization (RM) capture only one side of this trade-off: RSD is strategyproof but inefficient, while PS and RM are efficient but incentivize manipulation. 
We propose EMERGENT, a novel application of Generative Flow Networks (GFlowNets) to one-sided matching, leveraging its ability to sample diverse, high-reward solutions. 
In our approach, efficient and manipulation-resistant matches emerge naturally: 
high-reward solutions yield efficient matches, while the stochasticity of GFlowNets-based outputs reduces incentives for manipulation. 
Experiments show that EMERGENT outperforms RSD in rank efficiency while significantly reducing strategic vulnerability compared to matches produced by RM and PS. 
Our work highlights the potential of GFlowNets for applications involving social choice mechanisms, where it is crucial to balance efficiency and manipulability.

\end{abstract}

\begin{figure}[t]
    \centering
    \includegraphics[width=0.8\linewidth]{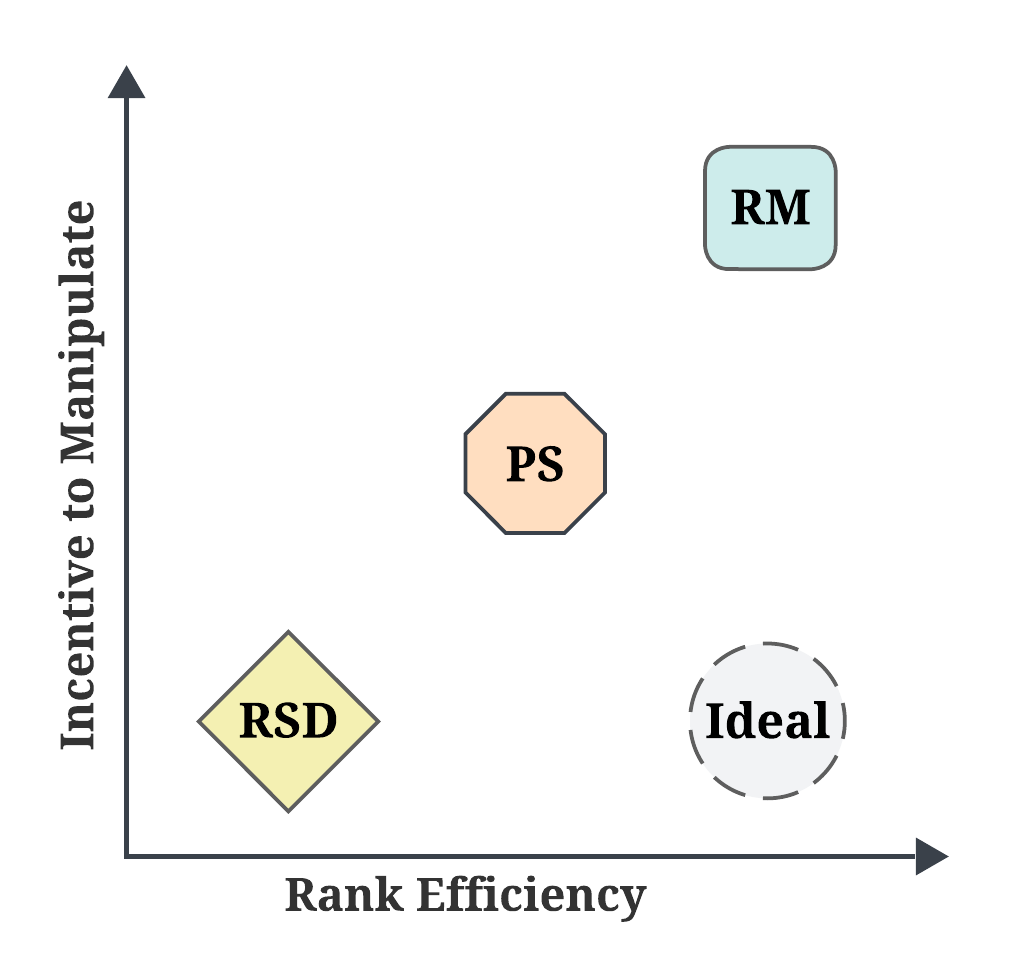}    
    \caption{The trade-offs of popular one-sided matching methods i.e., Random Serial Dictatorship (RSD), Probabilistic Serial (PS) and Rank Minimization (RM). A hypothetical "ideal" method provides maximum rank efficiency and no incentive to manipulate.}
    \label{fig:motivation}
\end{figure}

\begin{figure*}[t]
    \centering
    \includegraphics[width=0.9\linewidth]{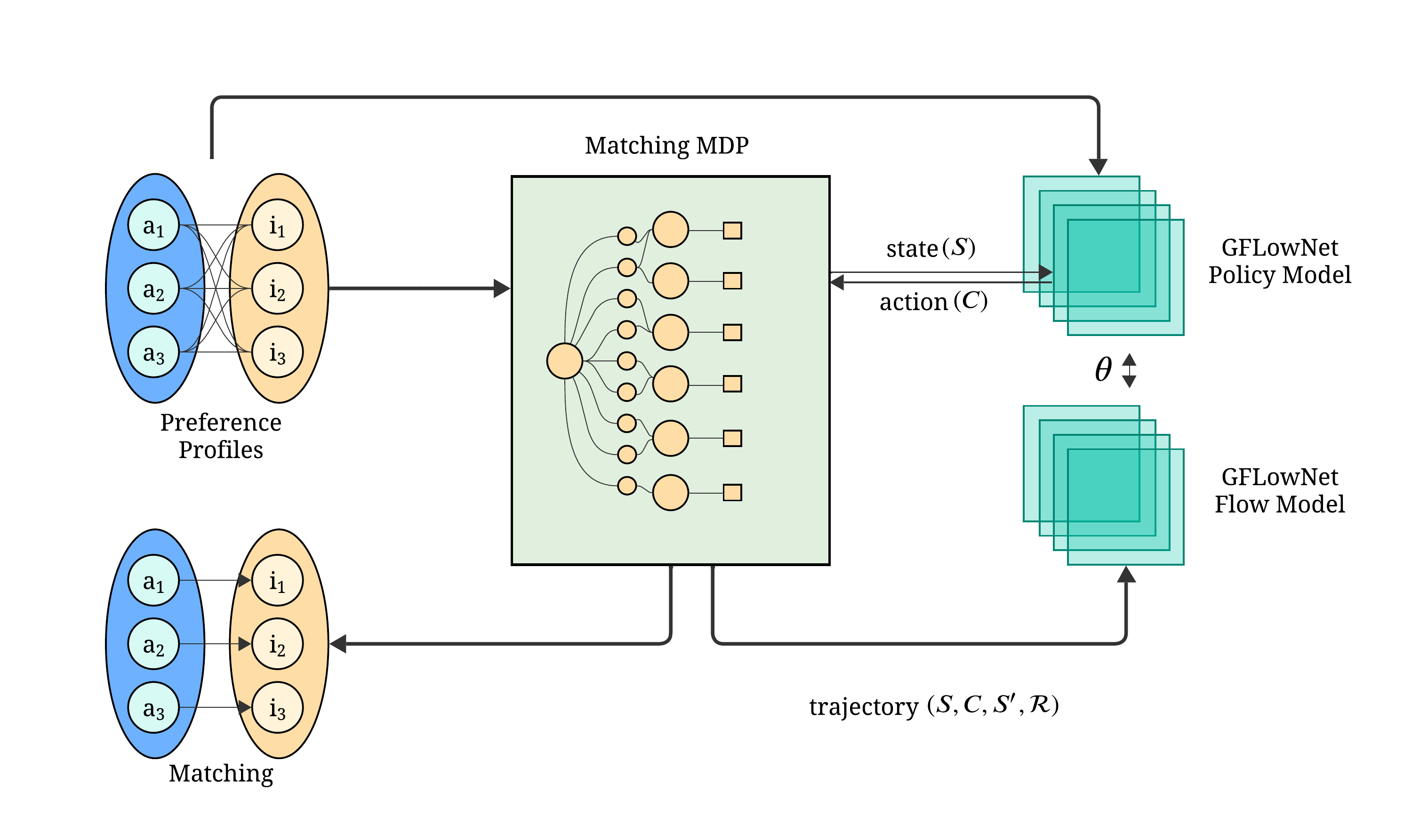}    
    \caption{\textbf{EMERGENT}. Training a GFlowNets-based model to create valid one-to-one matches over bipartite graphs.}
    \label{fig:architecture}
\end{figure*}

\section{Introduction}
\label{sec:intro}
Matching agents to items based on their preferences is a fundamental problem in economics, social choice theory, and machine learning, with applications ranging from school choice to resource allocation~\cite{abdulkadiroglu1998random, bogomolnaia2001new}. 
An ideal matching mechanism aims to balance two critical objectives: \textit{efficiency}, ensuring agents are matched to items they value highly, and \textit{strategyproofness}, ensuring agents have no incentive to misreport their preferences or manipulate the matching mechanism. 
However, it is well-established that no mechanism can simultaneously achieve both objectives~\cite{roth1982economics, svensson1999strategy}.

Existing mechanisms, such as Random Serial Dictatorship (RSD), Probabilistic Serial (PS), and Rank Minimization (RM), each address only one side of this trade-off. 
RSD is strategyproof but often produces inefficient matches, while PS and RM are more efficient but incentivize agents to manipulate their preferences~\cite{bogomolnaia2001new}. 
%

In this work, we propose \textbf{EMERGENT}, 
a novel approach based on Generative Flow Networks (GFlowNets), to one-sided matching problem. 
GFlowNets are generative models designed to sample diverse solutions in proportion to their rewards~\cite{bengio2021flow}. 
By framing the matching problem as a sequential decision process, EMERGENT learns to generate allocations that balance efficiency and manipulation resistance.~\footnote{We will use the term synonymously with \emph{strategyproofness}.} 
In particular, the diversity inherent in solutions generated by GFlowNets helps make the matching mechanism less vulnerable to manipulation, while providing consistently high, if not optimal, rank efficiency.
Notably, these properties \emph{emerge} from the generative process, without explicitly training the model to learn them.

We evaluate EMERGENT against established baselines on synthetic preference profiles for markets of varying sizes $( n = 3, 5, 7)$. 
Our results demonstrate that EMERGENT achieves higher rank efficiency than RSD and approaches the performance of PS, while significantly reducing strategic vulnerability compared to PS and RM. 
These findings suggest that EMERGENT is a promising new approach to tackling the efficiency-strategyproofness trade-off.

Our main contributions are as follows:
\begin{itemize}
    \item We introduce EMERGENT, a novel GFlowNet-based framework for one-sided matching.
    \item We show that EMERGENT produces matchings with high rank efficiency and low manipulation incentives.
    \item We introduce the Efficiency-Manipulation Trade-off (EMT) to characterize the distance between any matching method and an ideal rank efficient method which provides no incentive to manipulate.
    \item We provide experimental validation of EMERGENT's performance against baseline methods such as RSD, PS and RM.
\end{itemize}

\section{Preliminaries}
\label{sec:preli}
We formally define our problem setting, followed by some key concepts for measuring  efficiency and strategyproofness and the GFlowNets framework\cite{bengio2021flow}.

\subsection{Problem Setting}
\label{subsec:problem}

\noindent\textbf{One-sided matching.}  
Given a set $A = \{a_1, a_2, ..., a_n\}$  of $n$ agents  and a set $I = \{i_1, i_2, ..., i_n\}$ of $n$ items, each agent $a \in A$ has strict ranked preferences $R_a$ over the items in $I$.  
A \textit{matching mechanism}  $M : 2^R \rightarrow 2^{\{(a, i) \mid a \in A, i \in I \}}$ is a function that maps a given preference profile $R = (R_a)_{a \in A}$ to a set of pairs $(a, i)$ (\emph{matching}) such that each agent is \emph{matched} exactly to a single item and vice versa.~\footnote{A preference profile is also referred to as a \textit{market}.} 

In this work, we only consider complete and strict preferences.
The rank of an item $i$ for an agent $a$, denoted as $r_{(a,i)} \in \{1,2,\dots,n\}$, represents the position of $i$ in $a$'s preference list $R_a$.  
The objective of a \textit{rank-minimizing match} (RM) is to find a matching $M^*$ that minimizes the average rank:
\begin{equation*}
M^* = \arg\min_{M} \frac{1}{n} \sum_{(a, i) \in M} r_{(a,i)}.
\end{equation*}

\noindent\textbf{Rank Efficiency.}  
~\cite{featherstone2020rank} defines \textit{rank efficiency} as a refinement of \textit{ordinal efficiency}~\cite{bogomolnaia2001new}, where an assignment is rank efficient if its \textit{rank distribution} cannot be stochastically improved, and the RM method produces rank efficient matches.
Since verifying stochastic dominance is computationally intractable, we approximate \textit{rank efficiency (RE)} using the average rank of a given match $M$, $AR(M)={\frac{1}{n} \sum_{(a, i) \in M} r_{(a, i)}}$.
Consequently, we define rank efficiency as:
\begin{equation*}
   \text{RE}(M) = \frac{1}{\frac{1}{n} \sum_{(a, i) \in M} r_{(a, i)}}.
\end{equation*}
Higher values of RE($M$) indicate that agents, on average, are matched to items closer to the top of their preferences.\\

\noindent\textbf{Truthful \& Manipulated Preferences.}  
Let $u(a, i) \in \mathbb{R}$ denote agent $a$'s private utility for item $i$.  
An agent's preference list $R_a$ is said to be truthful if $\forall (i, j) \in I : r_{(a,i)} < r_{(a,j)} \iff u(a, i) > u(a, j)$.  
That is, the items in an agent's preference list are ranked in decreasing order of their private utility. Any deviation from this order is considered a \textit{manipulation} or \textit{misreport}.

\noindent\textbf{Strategyproofness.}  
A matching mechanism $M$ is said to be \textit{strategyproof} if no agent can achieve a strictly better outcome by misreporting their preferences~\cite{roth1982economics}.  
Let $R_a$ and $R'_a$ represent the true and misreported preference ordering of agent $a$, respectively. Then, $M$ is strategyproof if and only if:
\begin{equation*}
    u(a, M(R_a, R_{-a})) \geq u(a, M(R'_a, R_{-a})),
\end{equation*}
for all agents $a \in A$, all possible true preference profiles $R = (R_a, R_{-a})$, and all misreported preferences $R'_a$, where $R_{-a}$ denotes the preferences of all agents except $a$.  
%

\noindent\textbf{Measuring Resistance to Manipulation.}  
The degree to which a mechanism resists manipulation can be quantified using the \textit{incentive ratio}~\cite{chen2011profitable}. 
This metric measures the maximum improvement in utility that an agent can achieve by misreporting their preferences, compared to reporting truthfully.  
The incentive ratio for agent $a$ is defined as:
$$
\text{IR}_a = \frac{\max_{R'_a} u(a, M(R'_a, R_{-a}))}{u(a, M(R_a, R_{-a})}
$$
%
The overall incentive ratio for a mechanism $M$ is then computed as the average incentive ratio across all agents:
$$
\text{IR}(M) = \frac{1}{|A|} \sum_{a \in A} \text{IR}_a.
$$
An incentive ratio close to 1 indicates high resistance to manipulation, as agents gain little or no utility by misreporting their preferences.
%

\subsection{GFlowNets}
\label{sec:gfn}
Generative Flow Networks (GFlowNets) offer an approach to variational inference by framing the process of sampling from a target distribution as a sequential decision-making task~\cite{bengio2021flow}. 
We briefly summarize the formulation and the main training algorithms for GFlowNets used in this work.

Let there be a fully observable, deterministic MDP with a set of states $S$, a set of actions $C \subseteq S \times S$, and a designated \textit{initial state} $s_0$. 
The set of \textit{terminal states} is denoted by $X$. 
All states in the MDP are assumed to be reachable from the initial state $s_0$ via a sequence of actions.
A \textit{complete trajectory} is defined by a sequence of states $x = (s_0 \rightarrow s_1 \rightarrow \dots \rightarrow s_n)$, where $s_n \in X$ and every state pair is related via an action, i.e., $\forall i (s_i, s_{i+1}) \in C$.

The \textit{policy} in a GFlowNet, denoted as $P_F(s'|s)$, defines a probability distribution over the possible transitions from one state $s$ to a subsequent state $s'$. 
This policy induces a distribution over complete trajectories:
$$
    P_F (s_0 \rightarrow s_1 \rightarrow \dots \rightarrow s_n) = \prod_{i=0}^{n-1} P_F (s_{i+1} | s_i).
$$
The marginal distribution over terminal states, denoted by $P^\top_F(x)$, is induced by the policy, but computing it exactly can be intractable. 
%

The \textit{reward function} $R(x)=\exp(-\mathcal{E}(x)/T)$ assigns a positive value to terminal states and can be interpreted as an unnormalized probability mass over terminal states where $\mathcal{E}:X \rightarrow \mathbb{R}$ is an energy function and $T>0$ is a temperature parameter.
The GFlowNets objective is to learn a parameterized policy $P_F(s'|s;\theta)$ that generates a distribution $P_F^\top(x)$ over terminal states proportional to the reward function:

\begin{equation}
    P_F^\top(x) \propto R(x) = \exp\left(-\mathcal{E}(x)/T\right)
    \label{eq:reward}
\end{equation}
%
The forward policy $P_F(s' | s)$ is parameterized by a neural network with parameters $\theta$, which takes the state $s$ as input and produces the logits for transitioning to each possible next state $s'$. 
Learning this policy is challenging due to the intractability of computing the marginal distribution $P_F^\top$ induced by $P_F$ and the unknown normalization constant in Equation~\ref{eq:reward}. 
These difficulties are overcome by introducing auxiliary functions into the optimization process. 
In the following sections, we review two such objectives:

\noindent\textbf{Detailed Balance (DB).} 
The detailed balance (DB) objective \cite{bengio2023gflownet} requires learning two additional functions alongside the parametric forward policy $P_F(s' | s; \theta)$. 
These include a backward policy $P_B(s | s'; \theta)$, which defines a probability distribution over the predecessor states of any non-initial state in the MDP, and a state flow function $F(\cdot; \theta): S \to \mathbb{R}_{>0}$, which assigns positive flow values to states. 
The detailed balance loss for a transition $s \to s'$ is given by: 

\begin{equation*}
    \ell_{\text{DB}}(s, s'; \theta) = \left( \log \frac{F(s; \theta) P_F(s' | s; \theta)}{F(s'; \theta) P_B(s | s'; \theta)} \right)^2.
\end{equation*}

The DB training theorem states that if $\ell_{\text{DB}}(s, s'; \theta) = 0$ for all transitions $s \to s'$, then the forward policy $P_F$ successfully generates samples proportionally to the reward function. 

\noindent\textbf{Forward-Looking (FL).}  
The \textit{Forward-Looking (FL)} objective~\cite{pan2023better} incorporates intermediate learning signals to facilitate denser supervision. 
The FL loss is defined as:
\begin{equation*}
    \begin{split}
            \ell_{FL}(s, s'; \theta) = 
            \left( -\tilde{\mathcal{E}}(s) 
            + \log \tilde{F}(s; \theta) 
            + \log P_F(s' | s; \theta) 
            \right. 
            \\
            \left.
            + \tilde{\mathcal{E}}(s') 
            - \log \tilde{F}(s'; \theta) 
            - \log P_B(s | s'; \theta) \right){^2}
    \end{split}
\end{equation*}
where $\tilde{\mathcal{E}}(\cdot)$ extends the reward energy function $\mathcal{E}(\cdot)$ to non-terminal states, $\tilde{F}(s; \theta)$ is the approximate flow for these states. 
%
This approach enables faster credit assignment and improves training stability, as also reported by~\cite{zhang2023let}. 

%
%





\begin{figure*}[t]
    \centering
    \begin{subfigure}[t]{0.33\textwidth}
        \centering
        \includegraphics[width=1.0\textwidth]{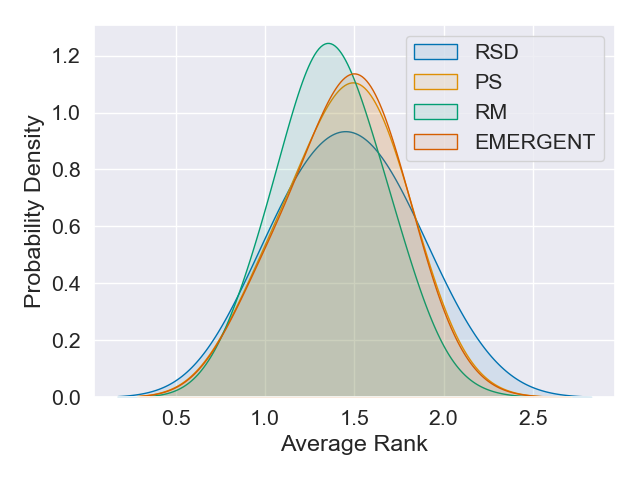}
    \end{subfigure}%
        \begin{subfigure}[t]{0.33\textwidth}
        \centering
        \includegraphics[width=1.0\textwidth]{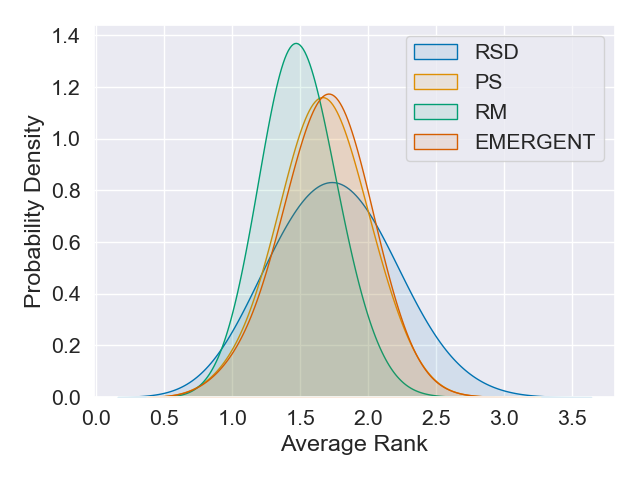}
    \end{subfigure}%
        \begin{subfigure}[t]{0.33\textwidth}
        \centering
        \includegraphics[width=1.0\textwidth]{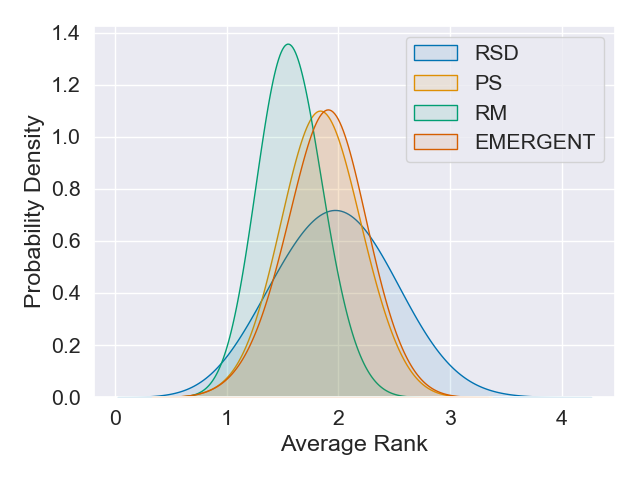}
    \end{subfigure}%
    \\
    \begin{subfigure}[t]{0.33\textwidth}
        \centering
        \includegraphics[width=1.0\textwidth]{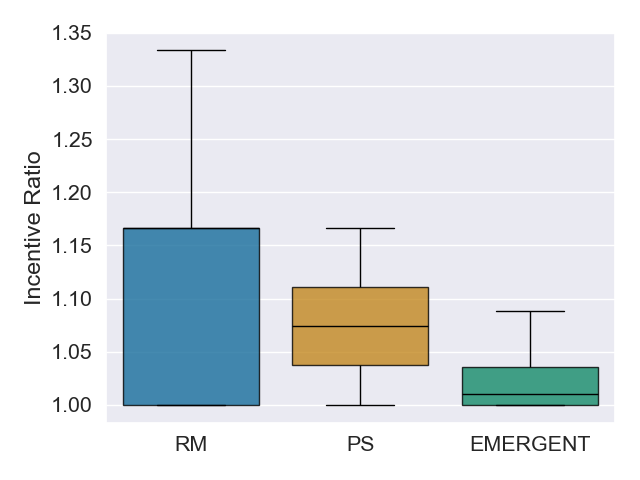}
        \caption{$n=3$}
    \end{subfigure}
        \begin{subfigure}[t]{0.33\textwidth}
        \centering
        \includegraphics[width=1.0\textwidth]{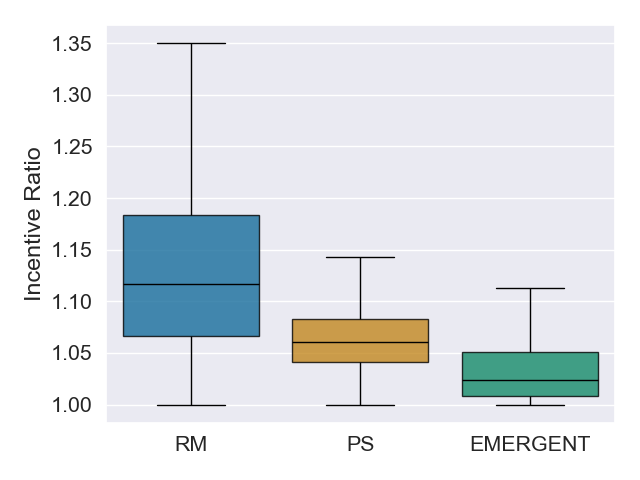}
        \caption{$n=5$}
    \end{subfigure}
        \begin{subfigure}[t]{0.33\textwidth}
        \centering
        \includegraphics[width=1.0\textwidth]{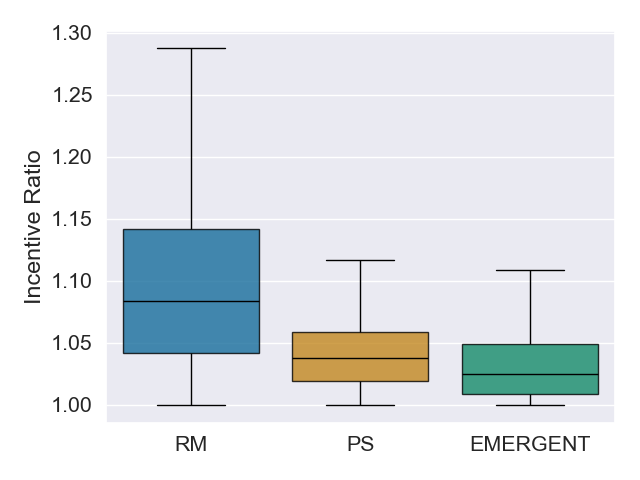}
        \caption{$n=7$}
    \end{subfigure}
    \caption{\textbf{Average Rank and Incentive Ratio.} Evaluating the efficiency and incentives given by matching algorithms. Matches made using GFN are more efficient and have lower average rank than Random Serial Dictatorship (RSD), while providing less incentive to manipulate than both the Hungarian and the Probabilistic Serial (PS) methods.}
    \label{fig:result}
\end{figure*}

\section{Method}
\label{sec:method}
We first characterize the Efficiency-Manipulation trade-off, by introducing EMT, metric that measures the distance of a given matching mechanism from an idealized method that maximizes rank efficiency and minimizes strategic incentives.

\subsection{Measuring the Efficiency-Manipulation trade-off.}
RSD has an incentive ratio of $1$ (perfect strategyproofness), while RM has the highest (most manipulable). 
They also have respectively the worst and the best rank efficiency.
To allow fair comparison, we normalize both metrics:
\begin{equation*}
\begin{split}
    \text{RE}_{\text{norm}}(M) = \frac{\text{RE}(M) - \text{RE}_{\text{RSD}}}{\text{RE}_{\text{RM}} - \text{RE}_{\text{RSD}}}, \\ \text{IR}_{\text{norm}}(M) = \frac{\text{IR}(M) - \text{IR}_\text{RSD}}{\text{IR}_{\text{RM}} - \text{IR}_\text{RSD}}.
    \end{split}
\end{equation*}
where $\text{RE}_{\text{RM}}$ is the highest rank efficiency (best case) and $\text{RE}_{\text{RSD}}$ is the lowest (worst case). 

An ideal matching mechanism would achieve both perfect efficiency ($\text{RE}_{\text{norm}}(M) = 1$) and perfect strategy resistance ($\text{IR}_{\text{norm}}(M) = 0$). 
However, such a mechanism is theoretically impossible. 

Given a mechanism $M$, EMT($M$) measures the Euclidean distance of a given method from this ideal point $(1,0)$:

\begin{equation*}
    \text{EMT}(M) = \sqrt{\left( \text{RE}_{\text{norm}}(M) - 1 \right)^2 + \left( \text{IR}_{\text{norm}}(M) - 0 \right)^2}.
\end{equation*}

A lower EMT value indicates a method that is closer to the ideal trade-off, meaning it efficiently assigns items while remaining strategy-resistant. 

\subsection{EMERGENT}
%
Extending the work of~\cite{zhang2023let}, we formulate one-sided matching as a Markov Decision Process (MDP) on a bipartite graph $G = (A \cup I, E)$, where edges $E$ encode the ranked preferences $R_a$ of agents over items.
Using this MDP, a GFlowNet policy model sequentially constructs matchings, guided by the FL objective.

Given $n$ agents and $n$ items, the matching MDP $(S,C,\mathcal{R})$ is defined as follows:\\

\noindent\textbf{States ($\mathbf{S}$).} A state $s \in S$ represents a partial matching and is denoted by a two-dimensional vector of size $(n \times n)$ where each element corresponds to a match between an agent and an item. 
The $(a,i)^{th}$ element of this vector denotes if the match between agent $a$ and item $i$ is a part of the final solution $X$.
Possible values ${0,1,2}$ represent 'does not belong to $X$', 'belongs to $X$', and 'not decided' respectively.

For example, a state $[[0,1,0],[1,0,0],[0,0,2]]$ signifies that agent 1 is matched with item 2, agent 2 is matched with item 1, and agent 3 hasn't been assigned yet. 
The initial state $s_0$ would have all agents unassigned. 
The terminal state space $X$ would consist of all states where every agent is assigned to a unique item. \\

\noindent\textbf{Actions ($\mathbf{C}$).} An action $c \in C$ consists of selecting agent $a$ and item $i$ such that $s[a,i]=2$ and after matching this agent-item pair the new state becomes $s'[a,i]=1$.
Each transition modifies the available actions in the policy.
For example, in $s'$, $a_j$ and $i_k$ can no longer be assigned to any other item or agent.
That is, $s'[a,:]=0$ and $s'[:,i]=0$.\\

\noindent\textbf{Reward ($\mathcal{R}$).} The reward function $\mathcal{R}(x)$ for a terminal state $x \in X$ is defined as the inverse of the sum of ranks achieved in that matching:
\begin{equation*}
    \mathcal{R}(x) = \frac{1}{1 + \sum_{(a, i) \in x} r_{(a,i)}} 
\end{equation*}
This reward ensures that matchings with lower total ranks have higher rewards. 
Note that we do not include incentive ratio into the reward, and despite this matches learned by our method exhibits less incentive ratio than baselines, as discussed later in Section~\ref{sec:results}.

The MDP is used to generate trajectory data through rollouts following the GFlowNets policy.  
Starting from the initial state $s_0$, the policy samples actions iteratively based on the current state and the learned probabilities of transitioning to subsequent states. 
This process continues until a terminal state is reached and a complete match is formed.  
Each rollout produces a trajectory $\tau$ consisting of a sequence of states and actions:

\begin{equation*}
\tau = (s_0, a_1, s_1, a_2, s_2, ..., a_t, s_t)
\end{equation*}

where $s_t$ is a terminal state and $t$ is the trajectory length.

\subsubsection{GFlowNets Training}
\label{subsec:training}

The collected trajectory data is used to train the GFlowNets model, which learns a policy parameterized by a Graph Isomorphism Network (GIN)~\cite{xu2018how}.  
Our architecture follows~\cite{zhang2023let}, utilizing separate GIN models sharing the same parameters $\theta$, for the forward policy and the state flow function to improve training stability.
Each GIN consists of five hidden layers with a hidden size of 256 dimensions. 
The input to the GIN is an integer vector representation of the state, where each element takes values from $\{0,1,2\}^{|A \times I|}$ and edge values $E$ encoding preference rankings, followed by an embedding layer. 
For the forward policy GIN, we set a unidimensional output node feature, followed by a softmax operation over all vertices to obtain a categorical distribution over matching actions. 
The state flow GIN applies an edge pooling layer to extract a unidimensional graph-level output that serves as the flow value.
The training process employs the Forward-Looking (FL) objective, which ensures that matchings are generated with probability proportional to their rewards. 
We minimize the FL loss function via stochastic gradient descent using trajectories $\tau \sim P_F(\tau; \theta)$ sampled from the forward policy.
Following~\cite{zhang2023let}, we fix the backward policy as a uniform distribution over all selected vertices for simplicity.  
We train our model using the Adam optimizer with a learning rate of $1 \times 10^{-3}$, without additional hyperparameter tuning.

The GIN model outputs a probability distribution over possible agent-item matches. 
This distribution is then used to sample the next action during rollouts, ensuring stochastic exploration.
To improve training efficiency, we break full trajectories into transition pieces, which are batched with a batch size of 64 and used to compute the FL loss.
Figure~\ref{fig:architecture} illustrates the full training pipeline of EMERGENT.



\section{Experiments}
\label{sec:experiments}
We conduct empirical experiments to show the effectiveness of EMERGENT in providing matches with high rank efficiency as well as low incentive to manipulate in different market settings.
We also show the flexibility of our method in navigating the efficiency-manipulability trade-off using the temperature parameter ($T$) of the energy function for GFlowNets.
Additionally, we assess the scalability of our method by training it on small markets and testing it on progressively larger markets.
Each data point is obtained by averaging across 100 different runs.

\subsection{Datasets}
\label{subsec:data}
We generate synthetic preference profiles for one-sided matching tasks, where $n$ agents have complete and strict ranked preferences over $n$ items, sampled uniformly at random.
The dataset includes profiles for markets of varying sizes $(n = 3,4,5,6,7)$ to evaluate the scalability and performance of our model across different problem complexities. 
Each profile is stored as a bipartite graph, where edges represent agent-item rankings, encoded as weights. 
The training and test data sets consist of 4,000 and 400 profiles for $n = 3, 4$, and 10,000 and 1000 for $n = 5, 6, 7$, respectively.
%

\subsection{Baselines}
\label{subsec:baselines}
We compare the performance of EMERGENT against three traditional algorithms for one-sided matching: 

\noindent\textbf{Random Serial Dictatorship (RSD):}  
RSD is a widely used matching mechanism that assigns items to agents based on a randomly generated priority order~\cite{abdulkadiroglu1998random}.
Agents sequentially select their most preferred available item according to this order, and therefore have no incentive to manipulate their preferences.

\noindent\textbf{Probabilistic Serial (PS):}  
PS is a fractional allocation mechanism in which agents simultaneously consume items at a uniform rate starting from their top-ranked preferences~\cite{bogomolnaia2001new}. 
This process continues until all items are fully consumed. 
PS is well-known for producing efficient matches with high welfare. 
However, the mechanism is not strategyproof, as agents may benefit from misreporting their preferences to gain more desirable allocations.

\noindent\textbf{Rank Minimization (RM):}  
In this mechanism, matches are done  with the sole principle of minimizing  the average rank of matched item.
RM is implemented using the classic Hungarian algorithm for solving the assignment problem~\cite{kuhn1955hungarian}. 
It provides rank-efficient matches with respect to agents’ ranked preferences. 
Despite this efficiency, the mechanism is vulnerable to manipulation from agents.


\begin{table*}[t]
\centering
 
\caption{Average rank (AR), Incentive Ratio (IR) and Efficiency-Manipulation trade-off (EMT) scores for EMERGENT and baselines.}
\label{tab:result}
\end{table*}

\subsection{Metrics}
\label{subsec:metrics}
We measure the rank-efficiency and manipulability of our method through \textit{average rank (AR)} and \textit{incentive ratio (IR)} respectively, previously defined in Section~\ref{subsec:problem}.
We also assess how closely all methods approximate the ideal matching method by computing their \textit{Efficiency-Manipulation Tradeoff (EMT)} scores.
Low AR indicates that agents receive items closer to the top of their preference lists.
%
%
%
Lower values of IR suggest that the mechanism incentivizes less strategic behavior.
The computation of the IR involves evaluating all possible misreports for each agent in the market. 
For a market with $n$ agents and $n$ items, the number of misreports grows factorially $(n \cdot n!)$, making the problem computationally intractable for even moderately sized markets. 
Therefore we approximate the computation of IR for $(n=5, 7)$ by evaluating a subset of agents and sampling possible misreported preferences uniformly for these agents.

\begin{figure}
    \centering
    \includegraphics[width=1.0\linewidth]{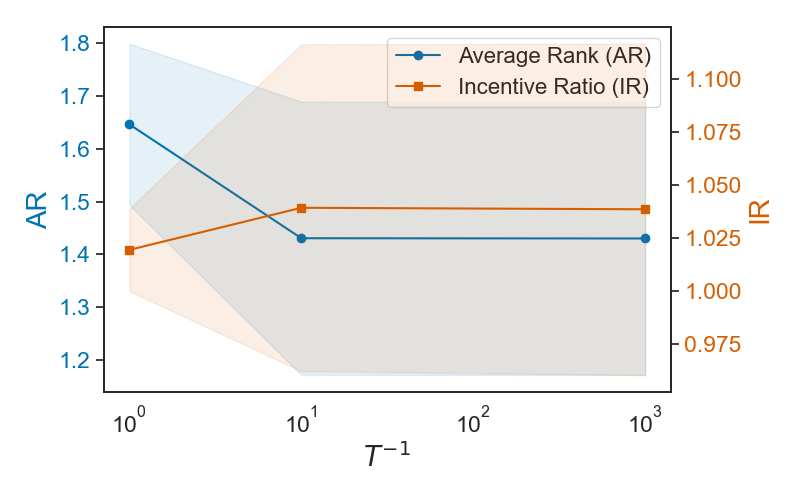}    \caption{\textbf{Flexibility.} Effect of the inverse temperature $T$ on rank efficiency and incentive ratio. The $x$-axis represents $T^{-1}$, with the $y$-axis showing average rank (left) and incentive ratio (right).}
    \label{fig:temperature}
\end{figure}

\section{Results}
\label{sec:results}
We evaluate the above metrics for each method, averaged over all preference profiles in the test dataset for markets of size $(n=3,5, 7)$.
Additionally each datapoint for EMERGENT is averaged over 100 runs.

We summarize our main findings below.

\subsection{Balancing Efficiency and Incentives}
Table~\ref{tab:result} and Figure~\ref{fig:result} present the average values of AR, IR and EMT for each method.
As expected, RSD exhibits the worst AR due to its random order of allocation, while RM achieves the best. 
Our method demonstrates AR nearly equivalent to PS and significantly better than RSD. 

%
RSD is strategyproof, and therefore exhibits an IR of 1.0.
In contrast, RM and PS show significantly higher IR.
Our method achieves lower incentive ratios compared to RM and even PS across all tested market sizes.

\begin{figure}
    \centering
    \includegraphics[width=0.9\linewidth]{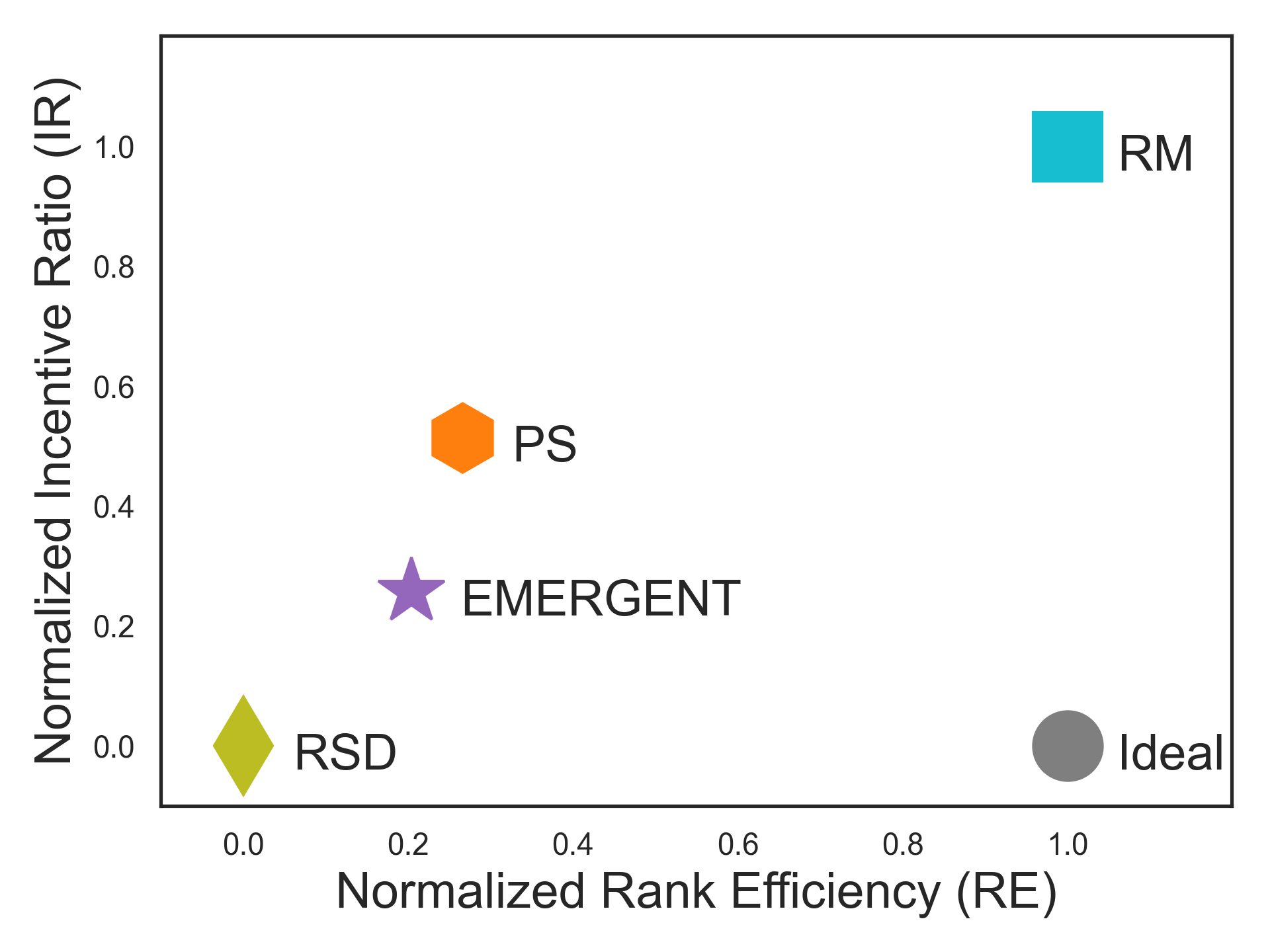}
    \caption{\textbf{Efficiency-Manipulation Trade-off.} Trade-off between RE ($x$-axis) and IR ($y$-axis) for all methods. EMERGENT is closest to the ideal matching method.}
    \label{fig:summary}
\end{figure}

EMERGENT also achieves the lowest EMT scores out of all tested methods for markets of size $n=3,5$, shown in Table~\ref{tab:result}.
Figure~\ref{fig:summary} illustrates this by showing the position of all methods with respect to the ideal matching method, with our method being the closest.
While PS and RM achieve higher rank efficiency (RE), they do so at the cost of a higher IR.
Our method, in contrast, attains a favorable balance, combining high RE with a significantly lower IR.

\begin{figure}
    \centering
    \includegraphics[width=1.0\linewidth]{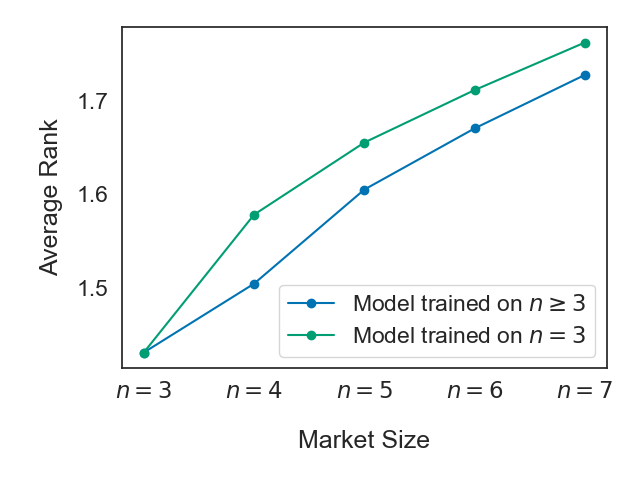}
    \caption{\textbf{Scalability.} The $y$-axis shows the gap in average rank across different market sizes $N$ for a model trained on $n=3$.}
    \label{fig:scale}
\end{figure}

\subsection{Flexibility}
We also evaluate the flexibility of our method when trained using different values for the temperature parameter $T$, as defined in Section~\ref{sec:gfn}.
As the reward for the GFLowNets policy is inversely proportional to AR, larger values of $T$ lead to a more uniform distribution over possible matches and therefore lead to higher AR.
Smaller values of $T$ lead to more rank-efficient matches with lower AR and consequently incentivize more manipulation.
We empirically show this in figure~\ref{fig:temperature}, where the $x$-axis shows inverse temperature $T^{-1}$, and a dual $y$-axis shows AR (left) and IR (right).
This shows the possibility of fine-tuning the performance of EMERGENT along the efficiency-manipulation trade-off.

\subsection{Scaling To Larger Markets}
\label{subsec:scalability}
To evaluate the generalization capabilities of our method, we conduct an experiment where a model trained on a small market ($n=3$) is tested on progressively larger markets. 
We compare its performance against models that are explicitly trained for each market size ($n=3,4,5,6,7$).

For each $n$, we measure the AR of matches for a model trained on $n=3$ and a model trained directly on the respective market size $n$. 
Our results, shown in Figure~\ref{fig:scale}, indicate that the gap in AR between the explicitly trained models and the $n=3$ model decreases as market size increases. 
Despite being trained only on a small market, the $n=3$ model continues to produce matches with low AR, approaching the performance of larger models.
We hypothesize that this occurs since the $n=3$ model benefits from complete exposure to all possible preference profiles (only 36 exist), whereas models trained on larger markets face an exponentially increasing preference space ($207M$ for $n=5$, $1.93 \times 10^{14}$ for $n=6$, and $1.64 \times 10^{22}$ for $n=7$). 
Given that the models are trained on 4,000-10,000 samples, larger models suffer from reduced preference coverage.
%
%
%
This suggests that EMERGENT captures structural patterns in matching markets, with the potential to generalize well even when trained on small markets.

\section{Discussion}
\label{sec:discussion}
Our findings highlight the potential of our approach,  EMERGENT, in addressing the fundamental trade-off between efficiency and incentive to manipulate in one-sided matching problems. 
Among all tested methods, EMERGENT achieves the lowest Efficiency-Manipulation Tradeoff (EMT) score, indicating that it provides the closest approximation to the ideal matching method. 
Another key strength of our approach is its ability to generalize across different market sizes. 
While the generalization gap increases with market size, the model continues to produce efficient matches, suggesting that it successfully learns structural properties of the matching problem. 
This property is particularly valuable in real-world applications where training data may be limited.

Our results also show that EMERGENT allows for flexible control over the trade-off between efficiency and strategic resistance through the temperature parameter $T$. 
By adjusting $T$, one can tune the distribution of generated matchings, ranging from nearly deterministic, highly efficient solutions to more randomized assignments that discourage manipulation. 
This flexibility is not present in standard mechanisms such as RSD or PS, where the trade-off is inherent to the method rather than a tunable parameter.

Despite empirical results, our method does not provide theoretical guarantees on incentives to manipulate. 
Future work could derive upper and lower bounds for incentive ratios for this method.
%
%
Future work should also explore other solution concepts in matching such as stability and envy-freeness~\cite{brandt2016handbook}.
Our current experiments focus on uniform preference distributions; it would be valuable to evaluate our method under real-world preference data, where strategic incentives may be more nuanced. 
Finally, scaling and evaluating EMERGENT to significantly larger markets ($n \gg 7$) remains an open challenge, particularly in terms of computational efficiency and training stability.



\section{Related Work}
\label{sec:related_work}
Matching algorithms have been extensively studied in economics and computer science, focusing on trade-offs between strategyproofness, fairness and efficiency. 
Random Serial Dictatorship (RSD), a popular traditional matching method, is strategyproof, ensuring that agents have no incentive to misreport preferences~\cite{abdulkadiroglu1998random}. 
However, RSD produces suboptimal matches in terms of rank efficiency~\cite{featherstone2020rank}, as it does not minimize the ranks of assigned items. 
Assuming $n$ agents and $n$ items, RSD provides matches that are ranked $\log(n)$ on average, growing larger with increasing market size~\cite{knuth1996exact, nikzad2022rank}.

In contrast, rank efficient methods such as Rank-Minimization (RM)~\cite{featherstone2020rank, aksoy2013school, troyan2022non, ortega2023cost} minimize the average rank received by all agents, but sacrifice strategyproofness, allowing agents to manipulate the matching mechanism~\cite{troyan2022non, aksoy2013school, tasnim2024strategic}. 
Similarly, the Probabilistic Serial (PS) mechanism provides a middle ground with ordinal efficiency~\cite{bogomolnaia2001new}, but remains vulnerable to manipulation~\cite{wang2020bounded}.

The efficiency-strategyproofness trade-off has been studied in the context of both matching and voting problems. 
~\cite{christodoulou2016social} analyzed the price of anarchy in one-sided matching, deriving theoretical welfare loss bounds due to strategic behavior. 
~\cite{anil2021learning} explored whether neural networks can learn welfare-maximizing voting rules while preserving incentives.

Several works have explored reinforcement learning (RL) solutions for dynamic matching problems.
~\cite{min2022learn} proposed an RL-based framework to provide stable matches that optimize social welfare in Markovian matching markets.
~\cite{liu2022welfare} examined RL applications to competitive equilibrium settings to maximize social welfare.

Inspired by RL, recent developments in machine learning has further enabled more adaptive generative approaches, such as Generative Flow Networks~\cite{bengio2021flow}, which sample from a distribution proportional to a reward function. In particular, 
~\cite{zhang2023let} demonstrated their efficacy in solving combinatorial optimization problems, but their potential for matching problems remains unexplored. 
Our work extends this paradigm by applying GFlowNets to one-sided matching.

\section{Conclusion}
\label{sec:conclusion}
We presented, EMERGENT, a GFlowNets-based model for generating rank-efficient matching with low incentives to manipulate preferences. 
The ability to generate rank efficient matchings that do not incentivize misreporting has significant implications for real-world applications such as school choice, resource allocation, and public housing assignments. 
Our approach navigates the trade-off between rank efficiency and incentive to manipulate, making it a promising approach for policymakers seeking allocation methods that can be fine-tuned for specific real-world contexts.
However, as with any data-driven approach, computational complexity and lack of interpretability will limit real-world applications.
Overall, our work demonstrates that GFlowNets offer a novel and effective paradigm for solving combinatorial problems, offering unexplored trade-offs.
We believe this approach opens new directions for future research in market design and incentive-aware resource allocation.




\section*{Impact Statement}

Beyond empirical contributions, our method has practical implications for policy design in education, housing, and labor markets, where improving both fairness and efficiency is critical. 
Additionally, our scalability experiments suggest that GFlowNets can be applied to larger real-world markets without the combinatorial explosion associated with traditional matching optimization methods. 
This work serves as a step towards AI-driven market design research, exploring deep generative models to provide novel solutions to long-standing economic and computational challenges. 
However, the lack of explainability and interoperability remains an unsolved issue, especially in the context of real-world applications.
Despite its advantages, our method relies on learned stochastic policies, which may introduce unpredictability.
AI and policy practitioners should therefore exercise caution when adopting this approach in a real-world context.

\nocite{langley00}





%


\bibliography{bibliography.bib}

\clearpage

\appendix

\end{document}